\documentclass[11pt]{article}

\usepackage[margin=1in]{geometry}
\usepackage{newtxtext,newtxmath}
\usepackage{booktabs}
\usepackage{graphicx}
\usepackage{amsmath}
\usepackage{multirow}
\usepackage{placeins}

\usepackage[numbers,sort&compress]{natbib}
\usepackage[colorlinks=true,linkcolor=blue,citecolor=blue,urlcolor=blue]{hyperref}
\usepackage{xcolor}

\newcommand{\keeprate}{\ensuremath{\rho}}

\title{Lost in Compression: A Controlled Cross-Lingual Audit\\of Extractive Prompt Compressors}

\author{
  Mantas Lukauskas\\
  Hostinger; Kaunas University of Technology\\
  \texttt{mantas.lukauskas@ktu.edu}
}

\date{\today}

\begin{document}
\maketitle

\begin{abstract}
Extractive prompt compression promises to cut LLM inference costs by removing
low-information tokens from the context, and learned compressors such as
LLMLingua-2 report strong results on English benchmarks. Most other
languages already pay a \emph{token premium}: the same content costs
1.3--1.8$\times$ more tokens than in English.
We ask whether compression closes or widens this gap. Using fully parallel
data in ten languages spanning five scripts, with controls budget-matched in
the target model's tokenizer, we audit four learned compressors against four
deterministic baselines, on eleven target models from ten vendors (over
250{,}000 evaluation calls). Three of the learned compressors are trained
with English supervision (LLMLingua-2 XLM-R/mBERT; Kompress-v2 from the
production Headroom stack); the fourth, XProvence, is a query-aware pruner
trained multilingually. We report three findings. First, the
transfer gap is real, replicates across target models and compressor
backbones, and is strongly rate-dependent: languages behave nearly alike at
a 0.75 keep-rate, but at 0.33 English retains 57--62\% of normalized
context utilization while Lithuanian retains 10--24\% and Chinese essentially
none, even though Chinese has the smallest token premium of the
nine languages. Second, the gap tracks compression supervision data,
not architecture. All three English-trained compressors show it (mean gap
$+0.23$ to $+0.31$ at keep-rate 0.5 on the primary model, up to 9 of 9
languages significant), deterministic methods show no comparable gap, and the
multilingually trained XProvence~v1 shows none at all. Its v2
release, retrained on translated data, empties 92\% of
Chinese contexts at its aggressive threshold without any warning. Third, in a harder long-context
setting, aggressive learned compression drives compressed contexts to or
below no-context utility in three of five non-English languages while
English remains usable. A
translate-then-compress pipeline matches or beats native compression at
roughly half the token cost in three of five tested languages. We release all
code, compressions, and model outputs. Safe compression budgets are
much smaller outside English.
\end{abstract}

\section{Introduction}
\label{sec:intro}

Long prompts dominate the cost of deploying LLM applications. Retrieved
documents, tool outputs, and conversation history are often orders of
magnitude longer than the question itself. \emph{Prompt compression} addresses this by
shortening the context before it reaches the target model. The most practical
family is extractive: a small model decides which tokens or sentences to keep,
the rest are dropped, and the compressed prompt remains plain text usable with
any API~\citep{jiang2023llmlingua,pan2024llmlingua2,li2023selective}.

These methods are developed, trained, and almost exclusively evaluated in
English. Yet the economics of tokenization already disadvantage other
languages: subword vocabularies allocate most of their capacity to English,
so lower-resourced languages and non-Latin scripts pay a persistent token
premium~\citep{petrov2023unfair,ahia2023cost}. On our parallel corpus, the same
passages cost 1.3--1.8$\times$ more o200k tokens in the nine non-English
languages than in English (Figure~\ref{fig:premium}), and up to 4.5$\times$
under Qwen2.5's tokenizer (Hindi). For this reason, compression has the
largest potential value in the languages where it has been tested the least.
If a compressor trained with English supervision degrades
disproportionately on non-English text, its users are penalized twice:
their tokens are more expensive, and the quality loss from compression is
larger.

This paper provides, to our knowledge, the first controlled cross-lingual audit
of extractive prompt compression. We keep the semantic content fixed (fully
parallel evaluation items in ten languages spanning five scripts), fix the
token budget in the \emph{target model's} tokenizer, and compare four learned
compressors (three English-supervised, one multilingually trained)
against deterministic baselines matched to the same achieved
budget, on eleven target models from ten vendors.

Our contributions:
\begin{enumerate}
  \item \textbf{A controlled audit protocol} that separates compressor quality
  from task difficulty, tokenizer effects, and model behavior, using
  paired parallel items, achieved-budget matching, and
  full/no-context/shuffled-context anchors (Section~\ref{sec:setup}).
  \item \textbf{Evidence that the transfer gap is rate-dependent and tied to
  compression supervision language} (Section~\ref{sec:results}): significant
  in 8 of 9 non-English languages at a 0.33 keep-rate (both primary target
  models), replicating across nine further target models and across all
  three English-supervised compressors (XLM-R, mBERT, ModernBERT backbones);
  deterministic methods show no comparable gap (isolated significant cases
  appear only for one model at the deepest budget), and neither does the
  multilingually trained XProvence~v1.
  \item \textbf{A long-context stress test} in which learned compression
  reduces non-English contexts to no-context performance
  (Section~\ref{sec:longctx}).
  \item \textbf{A translation-arbitrage analysis} showing
  translate-then-compress can dominate native compression
  (Section~\ref{sec:arbitrage}), with caveats about translationese.
\end{enumerate}

The rest of this paper is organized as follows.
Section~\ref{sec:related} reviews related work.
Section~\ref{sec:setup} describes the audit protocol: languages, conditions,
metrics, and target models. Section~\ref{sec:results} presents the main
results. Section~\ref{sec:longctx} reports the long-context stress test and
the task-dependence arm, and Section~\ref{sec:arbitrage} the translation
arbitrage analysis. Section~\ref{sec:discussion} discusses mechanisms,
practical recommendations, and limitations. Finally, conclusions are given
in Section~\ref{sec:conclusion}.

\section{Related Work}
\label{sec:related}

\paragraph{Prompt compression.} Perplexity-based token pruning
\citep{jiang2023llmlingua}, distilled token classification
\citep{pan2024llmlingua2}, query-aware reordering \citep{jiang2024longllmlingua},
and self-information filtering \citep{li2023selective} form the extractive
family we audit, together with production systems (Headroom's Kompress-v2
\citep{headroom2026}) and query-aware context pruners for RAG
\citep{chirkova2025provence,xprovence2026}; see \citet{li2025surveypc} for
a survey. Further public extractive systems (RECOMP \citep{xu2024recomp},
CPC \citep{liskavets2025cpc}, EXIT \citep{hwang2025exit}, and the
attention-probing Sentinel \citep{zhang2025sentinel}) all ship
English-supervised checkpoints (consistent with the pattern we document);
we exclude them from the audit because they are query-aware sentence
selectors at LLM scale or proxy-LM probes that do not fit our task-agnostic,
budget-matched protocol. A deployment-side study
\citep{kummer2026wild} measures LLMLingua-family latency and \emph{rate
adherence} at scale, in English; we find the same adherence failures to
be strongly language-dependent. A complementary line compresses into
continuous
representations (gist tokens \citep{mu2023gist}, in-context autoencoding
\citep{ge2024icae}, and KV-cache eviction or quantization
\citep{zhang2023h2o}), but these require access to model internals and are
unusable with commercial APIs, so we restrict the audit to text-to-text
extractive methods. Transferability audits of LLMLingua-2 exist along other
axes, e.g., to diffusion LLM targets \citep{huang2026dllm}, where compression
failures were likewise traced to omission of task-critical information; the
cross-lingual axis has, to our knowledge, not been audited.

\paragraph{Tokenizer inequality.} \citet{petrov2023unfair} and
\citet{ahia2023cost} established that subword tokenizers price non-English
text higher. We treat the premium as measured context (not a contribution)
and study how it interacts with compression.

\paragraph{Multilingual compression.} Almost all extractive compressors are
built and tuned on English: LLMLingua-2 distills GPT-4 judgments on
(English) MeetingBank, and Provence \citep{chirkova2025provence} trains on
English MS~MARCO. The recent exception is XProvence
\citep{xprovence2026}, which extends Provence multilingually over a BGE-M3
reranker; the v1 checkpoint we audit is trained on MS~MARCO and MIRACL
with silver sentence labels from a multilingual LLM (aya-expanse-8b) in 16
languages (a later v2 trains on translated MS~MARCO instead; we audit
both). It reports
strong multilingual pruning but does not measure a cross-lingual
\emph{transfer gap} relative to English on a shared target model, which is
our object of study. We include XProvence as the one multilingually
supervised point in our audit, and find that it is the only one that closes
the gap. Concurrently, a Chinese community fork of Headroom
(\texttt{headroom-zh}) added a dedicated Chinese lane because the upstream
English compressor was a no-op on Han script, and upstream has since merged
CJK-aware segmentation into its non-learned text lane (the learned
Kompress-v2 lane we audit remains whitespace-based as of v0.32). This is
independent, deployment-side confirmation of the failure mode we quantify.

\paragraph{Multilingual evaluation.} Belebele \citep{bandarkar2023belebele}
provides parallel reading-comprehension MCQ in 122 languages;
MultiEURLEX \citep{chalkidis2021multieurlex} parallel legal documents with
EUROVOC labels. LLMLingua-2's own evaluation is English-only despite its
multilingual XLM-R backbone \citep{conneau2020xlmr}; its training supervision
(GPT-4 distillation on MeetingBank) is entirely English. Our audit tests
whether the backbone's multilingual pretraining is enough for the
compression skill to transfer. We find that it is not.

\section{Audit Protocol}
\label{sec:setup}

\paragraph{Languages and data.} Ten languages with full parallelism:
EN, PL, FI, ET, LV, LT, UK, ZH (Simplified), AR (Modern Standard), HI.
These cover four language families (Indo-European, Uralic, Sino-Tibetan,
Afro-Asiatic) spanning seven branches (Germanic, Slavic, Baltic, Finnic,
Sinitic, Semitic, Indo-Aryan) and five scripts (Latin, Cyrillic, Han,
Arabic, Devanagari), with o200k token premiums from 1.28 (ZH) to 1.83 (LV). Primary task: Belebele reading
comprehension (300 parallel items per language). Long-document task:
MultiEURLEX level-1 EUROVOC classification (24 parallel documents).
Long-context task: target passage embedded among 7 same-language
distractors ($\sim$2--3k tokens; 150 items); these two auxiliary arms use
the six core European languages.

\paragraph{Conditions.} For each item: \texttt{full} (uncompressed),
\texttt{no\_context} (contamination/prior anchor), LLMLingua-2
\citep{pan2024llmlingua2} at requested keep-rates
$\keeprate \in \{0.75, 0.5, 0.33\}$, TF-IDF sentence extraction at the same
requested rates, deterministic lemmatization+stopword removal (its natural
budget is measured, not controlled), and two controls matched to LLMLingua-2's
\emph{achieved} o200k budget per item: prefix truncation and seeded random
word deletion. LLMLingua-1 \citep{jiang2023llmlingua} was excluded after
failing a technical gate: it does not compress Lithuanian at all (achieved
rate 0.99), which is itself evidence that perplexity-based pruning with a
small proxy LM does not transfer reliably. We also audit two more
learned compressors: Kompress-v2 \citep{headroom2026}, the ModernBERT prose
compressor inside the widely deployed Headroom production layer (trained
with English-dominant supervision spanning 17 text domains, including
agentic traces and meeting transcripts; on Chinese its learned lane is
effectively a no-op, returning text nearly unchanged at every requested
ratio (achieved keep-rate 0.91--0.96), a
failure independently confirmed by the community \texttt{headroom-zh} fork,
and upstream has since added CJK-aware segmentation to its \emph{non-learned}
text lane, while the Kompress-v2 lane we audit is unchanged as of v0.32),
and XProvence \citep{chirkova2025provence,xprovence2026}, a query-aware
multilingual context pruner built on a BGE-M3 reranker and trained on 16
languages with silver labels from a multilingual LLM, reaching 100+
languages via the backbone's cross-lingual transfer (released
CC~BY-NC-ND~4.0, research use only). XProvence is the one compressor in
our set designed for multilingual use, which lets us test whether
multilingual compression training closes the gap. We audit both released
checkpoints: v1 (natively multilingual MIRACL training data) and v2
(retrained on translated MS~MARCO), which turn out to behave very
differently on Chinese (Section~\ref{sec:results}).

\paragraph{Metric.} Accuracy is normalized against both anchors:
\begin{equation}
  U_{\ell,c} \;=\;
  \frac{\mathrm{acc}_{\ell,c} - \mathrm{acc}_{\ell,\mathrm{no\text{-}context}}}
       {\mathrm{acc}_{\ell,\mathrm{full}} - \mathrm{acc}_{\ell,\mathrm{no\text{-}context}}},
\end{equation}
the fraction of usable context value retained under condition $c$ in language
$\ell$. The \emph{transfer gap} is $\tau_{\ell,c} = U_{\mathrm{EN},c} - U_{\ell,c}$.
The denominator (full minus no-context accuracy) is comparable across
languages (25.7--31.7~pp on the primary model), so normalization does not
systematically amplify noise for any particular language.
Uncertainty: item-level paired bootstrap (2{,}000 resamples), identical items
across languages. We report per-comparison 95\% CIs without multiplicity
correction; across languages, methods, rates, and models this implies
several hundred tests, so isolated significances should be discounted.
We therefore emphasize patterns that replicate across models, methods,
and rates rather than any single interval.

\paragraph{Target models.} Compression is a cost-reduction technique, so we
audit the cost-efficient deployment tier (mini/flash/lite class) where it
is economically relevant; by production token volume, this is also
where most real traffic runs. Primary target models: \texttt{gpt-5.4-mini}
and \texttt{claude-haiku-4-5} (full grid, $n=300$). Replication targets on
a reduced grid ($n=150$; \texttt{full}, \texttt{no\_context}, LLMLingua-2,
TF-IDF, truncation at $\keeprate \in \{0.5, 0.33\}$): the newer-generation
\texttt{gpt-5.6-luna}, \texttt{gemini-3.5-flash}, five open-weights models
served via a gateway (Llama~4~Maverick, Mistral~Medium~3.5,
DeepSeek~V4~Flash, Kimi~K2.6, MiniMax~M3), and two proprietary
gateway-served models: Amazon's Nova~2~Lite as a price-floor point and
Alibaba's Qwen~3.7~Plus, covering the most multilingually marketed model
family. The four Chinese-vendor models
(DeepSeek, Kimi, MiniMax, Qwen) are included deliberately, to test whether
heavy
Chinese pretraining protects against the Chinese-language failure we
observe (we find no protective effect, Section~\ref{sec:results}); the nine-vendor roster
was fixed before any replication data were analyzed, and Qwen~3.7~Plus was
added once during revision, before its data were seen. Temperature 0;
instruction
language held constant (English) with passages, questions, and options in
the item language.

\section{Results: Transfer Gap by Rate, Method, and Target Model}
\label{sec:results}

\begin{figure}[t]
  \centering
  \includegraphics[width=\textwidth]{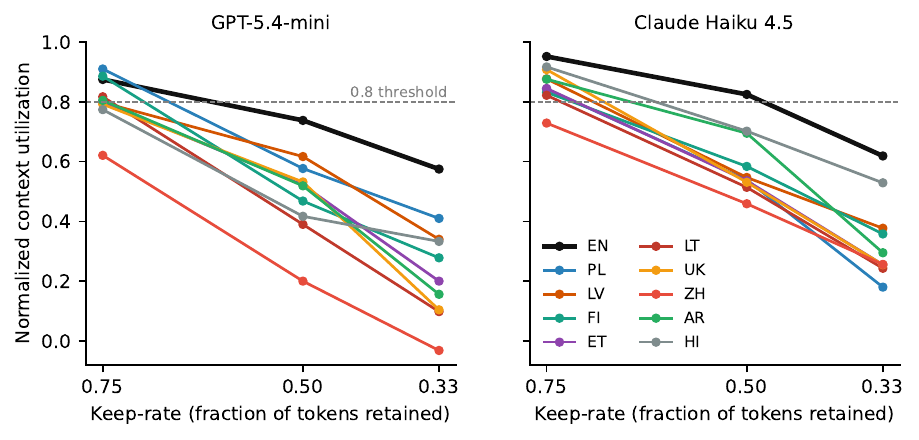}
  \caption{Normalized context utilization of LLMLingua-2 compression by
  keep-rate. English (black) declines slowly; the nine non-English
  languages drop sharply between $\keeprate=0.75$ and $\keeprate=0.33$. The
  pattern replicates across both target models.}
  \label{fig:rate}
\end{figure}

\begin{table}[t]
\centering\small
\caption{Normalized context utilization $U$ for LLMLingua-2
(GPT-5.4-mini / Claude Haiku 4.5), $n=300$ paired items. Bold: transfer gap
vs.\ EN significant (95\% bootstrap CI excludes 0).}
\label{tab:main}
\begin{tabular}{lccc}
\toprule
Language & $\keeprate=0.75$ & $\keeprate=0.5$ & $\keeprate=0.33$ \\
\midrule
EN & 0.88 / 0.95 & 0.74 / 0.82 & 0.57 / 0.62 \\
PL & 0.91 / \textbf{0.84} & 0.58 / \textbf{0.54} & 0.41 / \textbf{0.18} \\
LV & 0.80 / 0.88 & 0.62 / \textbf{0.55} & \textbf{0.34} / \textbf{0.38} \\
FI & 0.89 / \textbf{0.83} & \textbf{0.47} / \textbf{0.58} & \textbf{0.28} / \textbf{0.36} \\
ET & 0.80 / \textbf{0.84} & \textbf{0.52} / \textbf{0.54} & \textbf{0.20} / \textbf{0.26} \\
LT & 0.82 / \textbf{0.82} & \textbf{0.39} / \textbf{0.51} & \textbf{0.10} / \textbf{0.24} \\
UK & 0.79 / 0.91 & \textbf{0.53} / \textbf{0.53} & \textbf{0.10} / \textbf{0.26} \\
ZH & \textbf{0.62} / \textbf{0.73} & \textbf{0.20} / \textbf{0.46} & \textbf{$-$0.03} / \textbf{0.26} \\
AR & 0.81 / 0.88 & \textbf{0.52} / \textbf{0.69} & \textbf{0.16} / \textbf{0.29} \\
HI & 0.77 / 0.92 & \textbf{0.42} / \textbf{0.70} & \textbf{0.33} / 0.53 \\
\bottomrule
\end{tabular}
\end{table}

At mild compression ($\keeprate = 0.75$) languages are near-indistinguishable
for GPT (only Chinese shows a significant gap); at $\keeprate = 0.33$ the gap
is significant in 8/9 non-English languages for both target models
(Table~\ref{tab:main}, Figure~\ref{fig:rate}). English retains a majority of
the context value at every tested rate. Lithuanian and Ukrainian at
$\keeprate = 0.33$ retain 10\% (GPT), and Chinese drops below the
no-context anchor ($U = -0.03$), which means the compressed context is
worse for the model than no context. Chinese also has the smallest token premium in our
set (1.28$\times$ under o200k) yet the largest compression penalty, so
tokenizer pricing and compression safety are decoupled. In practical terms,
the \emph{safe compression budget} (deepest rate with $U \ge 0.8$) is about
2$\times$ compression for English but only $\sim$1.3$\times$ or less
for the other nine languages.

\begin{figure}[t]
  \centering
  \includegraphics[width=0.72\textwidth]{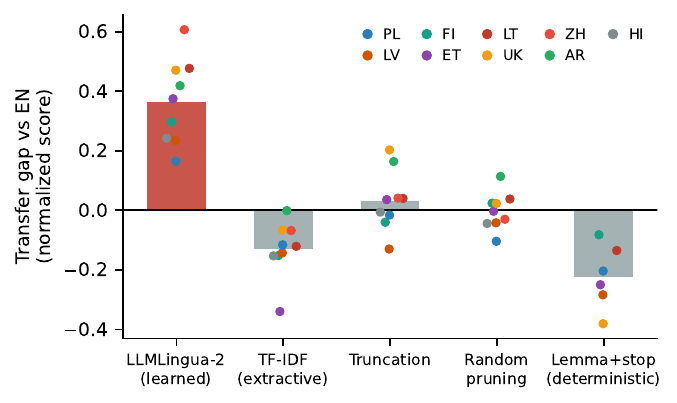}
  \caption{Mean transfer gap vs.\ EN at $\keeprate=0.33$ (GPT-5.4-mini; dots =
  languages). Only the learned compressor exhibits a cross-lingual gap on
  GPT; budget-matched deterministic methods do not (see text for the partial
  exception on Claude at this rate). This suggests the gap comes from the
  compressor rather than from the task, tokenizer, or target model.}
  \label{fig:methods}
\end{figure}

\paragraph{Deterministic baselines show no comparable gap.}
Figure~\ref{fig:methods} shows the key control: at the same achieved
budgets, TF-IDF extraction, truncation, random pruning, and
lemmatization+stopword removal show no comparable cross-lingual gap. On
GPT-5.4-mini no deterministic method reaches significance in any of the
nine languages at any rate (means $-0.20$ to $-0.09$ at $\keeprate=0.5$),
versus $+0.27$ (7/9 significant) for LLMLingua-2. On Claude the
deterministic means at $\keeprate=0.5$ are $-0.04$ to $+0.13$ (at most 2/9
significant) versus $+0.26$ (9/9) for LLMLingua-2; at the deepest budget
($\keeprate=0.33$) truncation and random deletion do develop gaps on
Claude (5/9 and 6/9 significant, means $+0.16$/$+0.19$), but with a
language profile unlike the learned compressors': Chinese, the worst
language under every learned method, is among the least affected
($+0.04$--$+0.06$). This is consistent with Claude-specific sensitivity to
heavily degraded input (see the refusal analysis below) rather than with
selection bias. The consequence for method rankings is large. In English
at $\keeprate = 0.33$, LLMLingua-2 is 12--14 accuracy points ahead of
TF-IDF. Across the nine non-English languages this advantage shrinks to
between $-5$ and $+7$ points, and the ranking reverses in ET/LT/ZH on GPT
and in FI/ET/LT/ZH on Claude. Therefore, an English-only evaluation would
rank these methods in the wrong order for a large share of our languages.

\paragraph{Replication with an mBERT backbone.} To rule out an XLM-R
idiosyncrasy, we repeat the audit with the mBERT variant of LLMLingua-2
(same English MeetingBank supervision, different multilingual encoder). The
pattern replicates: at $\keeprate = 0.5$ the mBERT compressor shows
significant gaps in 7/8 non-English languages on GPT-5.4-mini (mean $+0.31$,
vs.\ $+0.23$ in 6/8 for XLM-R; ZH excluded, see below), and the two
backbones' per-language gaps correlate.
Whatever multilingual knowledge the encoders carry from pretraining, the
English-only compression supervision does not exploit it. The mBERT variant
also fails \emph{rate adherence} on Chinese: requesting
$\keeprate=0.75$ yields an achieved o200k rate of 1.20, so the ``compressed''
text is longer than the original, because mBERT's character-level
WordPiece segmentation of Han script breaks the token-budget accounting. We
therefore exclude ZH from the mBERT gap statistics above; budget control
itself is a transfer failure mode that English-only evaluations cannot
detect. The XLM-R variant fails in the opposite direction on Chinese,
over-compressing (achieved 0.247 at requested 0.33, vs.\ 0.307 for
English). Our truncation/random controls are matched to these achieved
budgets, so the method-vs-control comparison remains fair; but the
ZH-vs-EN comparison compounds selection quality with budget control,
both of which are transfer failures of the same compressor.

\begin{figure}[t]
  \centering
  \includegraphics[width=0.72\textwidth]{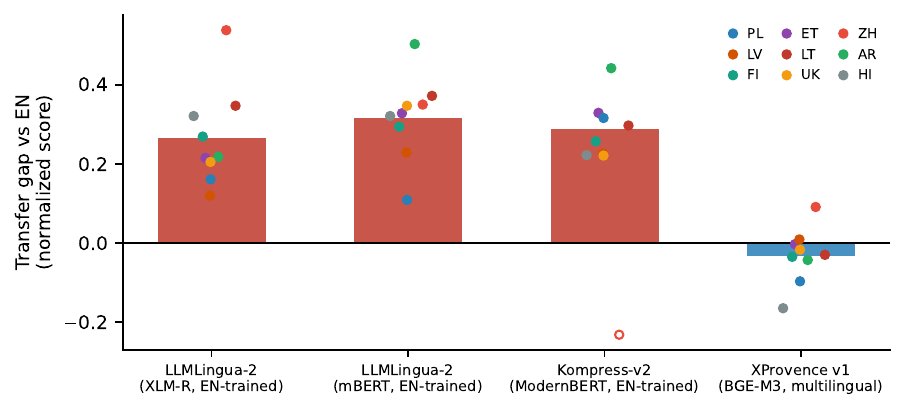}
  \caption{Transfer gap vs.\ EN at keep-rate 0.5 for the four learned
  compressors (GPT-5.4-mini, $n=300$). The three English-supervised
  compressors show large positive gaps regardless of backbone; the
  multilingually trained XProvence shows none. Kompress ZH (open marker) is
  excluded from its bar: Kompress is a near no-op on Han script (achieved
  rate 0.96), so its ``compressed'' Chinese is essentially the full context.}
  \label{fig:learned}
\end{figure}

\paragraph{Supervision language, not architecture, drives the gap.}
Figure~\ref{fig:learned} completes the compressor audit with two further
learned systems. Kompress-v2, a ModernBERT token classifier trained
with English-dominant, 17-domain supervision and deployed inside the
Headroom production stack,
reproduces the LLMLingua-2 pattern almost exactly: at its achieved
$\keeprate \approx 0.5$ budgets the gap is significant in 8/8 auditable
non-English languages on both primary target models (mean $+0.29$ GPT,
$+0.39$ Claude). It also inherits both transfer
failure modes seen before, in amplified form: weak rate adherence outside
English (requesting 0.33 achieves only 0.55--0.60) combined with barely
compressing English at all (requesting 0.5 keeps 0.89 of English vs.\
0.71--0.79 elsewhere; its small English penalty therefore partly reflects
that English text is left nearly intact, and its gap is a joint failure of
selection quality and budget control), and a structural no-op on Chinese
(achieved rate 0.91--0.96; whitespace tokenization cannot segment Han
script, see Section~\ref{sec:setup}). Unlike the LLMLingua-2 grid, the
Kompress and XProvence arms carry no re-matched truncation/random controls
at their achieved budgets, so we read their per-language levels
qualitatively and lean on the within-compressor EN-vs-rest contrast.
By contrast, XProvence (a
query-aware pruner built on the multilingual BGE-M3 reranker and trained on
16 languages) shows no transfer gap on GPT-5.4-mini at either
pruning threshold
(mean $-0.03$ at threshold 0.5 with achieved rates of 0.47--0.67; no
language significant, $n=300$), and the null replicates on Claude at
threshold 0.1 (mean $+0.005$, 0/9 significant). At Claude's aggressive
threshold-0.5 setting two languages reach significance (ET $+0.14$; ZH
$+0.38$), but the Chinese effect is dominated by Claude's refusal behavior
(it abstains on 25.7\% of pruned-Chinese calls, scored as incorrect)
rather than by role-confusable extraction errors. XProvence operates in an
easier,
query-aware setting than the task-agnostic compressors, so absolute scores
are not directly comparable; the cross-lingual contrast, however, is:
with multilingual supervision the EN-vs-rest asymmetry essentially
disappears. Two points could be read as caveats, but both of them support
the null result. First,
XProvence's achieved keep-rates are higher outside English (ZH 0.67,
HI 0.83 vs.\ EN 0.47 at threshold 0.5): part of its cross-lingual safety is
calibrated conservatism (pruning less where it is less certain); the
English-supervised compressors lack this behavior. Second,
only four of our nine non-English languages (FI, ZH, AR, HI) are among its
16 training languages; PL, ET, LV, LT, and UK are reached only through the
multilingual backbone's cross-lingual transfer, and show no gap either.

The v2 release of XProvence, retrained on translated MS~MARCO
rather than the natively multilingual MIRACL data behind v1, strengthens
the supervision-data point. At its conservative threshold (0.1) v2
replicates the v1 null on both primary models (mean $-0.09$ GPT / $+0.01$
Claude, 0/9 significant). At the aggressive threshold (0.5), however, a
gap re-appears (GPT $+0.16$, 3/9 significant; Claude $+0.29$, 7/9), and
its source is a Chinese calibration failure of a new kind: v2's relevance
scores for Chinese are degenerate, keeping essentially the whole passage
at threshold 0.1 (achieved rate 1.02) yet deleting everything
(returning an empty context) for 92\% of Chinese items at threshold 0.5
(achieved rate 0.08, vs.\ 0.21--0.24 in the other languages; ZH gap up to
$+0.97$ on Claude). Multilingual supervision closes the gap only insofar
as its scores are calibrated per language, and translated training data
does not guarantee that. A segmentation artifact compounds the Chinese
failure and clarifies its interpretation: the multilingual sentence splitter
XProvence ships with treats each Chinese passage as a single unit (mean 1.0
segments per passage, vs.\ 3.4--4.3 in the other scripts), so Chinese
pruning is all-or-nothing for both checkpoints: intermediate
keep-rates occur on $\le$1\% of Chinese items, against 42--77\% for
English. The v1--v2 contrast on Chinese is therefore pure score
calibration (v1's relevance scores clear the threshold, with 65\% of
passages kept whole at threshold 0.5, while v2's fall below it, with 92\%
deleted), and fine-grained Chinese pruning is not achievable with XProvence
as shipped. Therefore, the gap is not caused by learned compression as
such, and it is not caused by any particular encoder. It is inherited from
the compression supervision data. English-only supervision produces the gap
everywhere. Natively multilingual supervision (XProvence~v1) removes it.
Translation-based supervision (v2) removes it only at conservative
operating points.

\begin{figure}[!tp]
  \centering
  \includegraphics[width=0.9\textwidth]{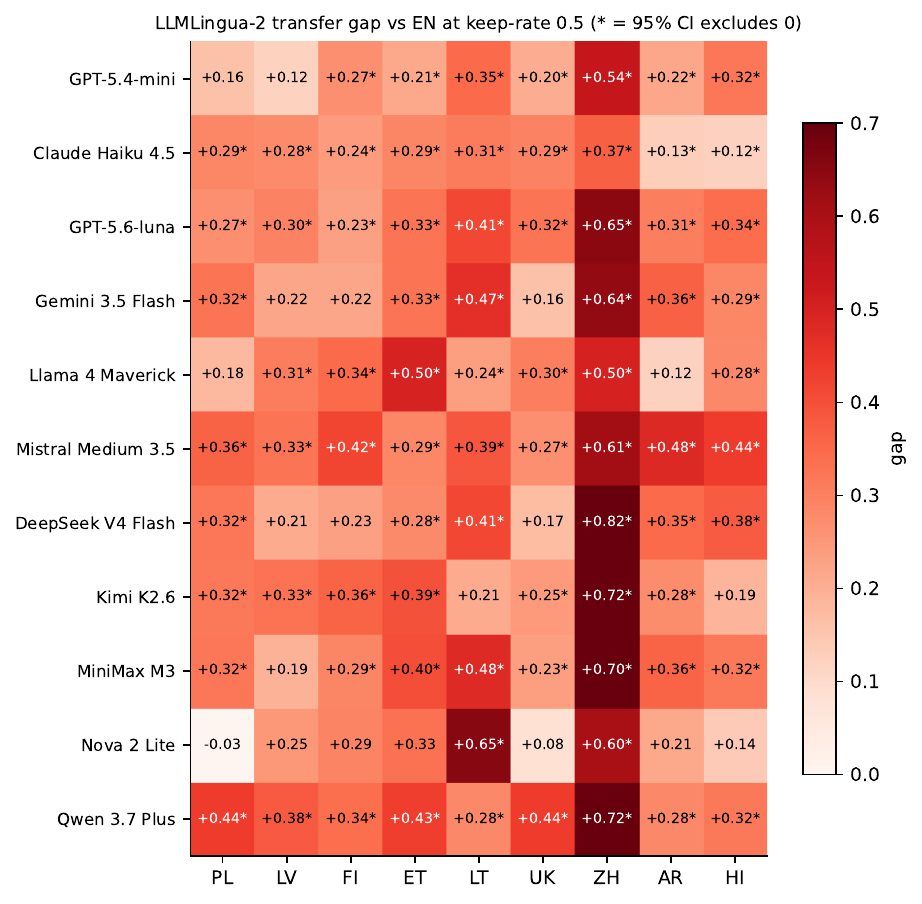}
  \caption{LLMLingua-2 transfer gap vs.\ EN at $\keeprate=0.5$ across target
  models ($*$: 95\% bootstrap CI excludes 0; additional models use $n=150$
  items and a reduced condition grid). The gap replicates across vendors and
  model generations; Chinese is consistently hit hardest.}
  \label{fig:models}
\end{figure}

\paragraph{Replication across eleven target models.} Figure~\ref{fig:models}
extends the audit to nine further target models on a reduced grid:
GPT-5.6-luna,
Gemini~3.5~Flash, five open-weights models served via a gateway
(Llama~4~Maverick, Mistral~Medium~3.5, DeepSeek~V4~Flash, Kimi~K2.6,
MiniMax~M3), and the proprietary Nova~2~Lite (Amazon) and Qwen~3.7~Plus
(Alibaba). Every model shows the same qualitative ordering (EN safest, ZH
consistently hardest), with mean gaps at $\keeprate=0.5$ of $+0.26$ to
$+0.40$ (per-language gaps up to $+0.82$; per-model numbers with significance
counts in Table~\ref{tab:allmodels}). The newer-generation
GPT-5.6-luna handles
compressed non-English input no better than the older mini model (mean gap
$+0.35$ vs.\ $+0.27$). A newer target model cannot recover the information
that the compressor has already deleted. The
four Chinese-vendor models (DeepSeek, Kimi, MiniMax, Qwen), despite far
heavier
Chinese pretraining, suffer the largest Chinese gaps of all
($+0.82$, $+0.72$, $+0.70$, $+0.72$), so the failure sits in the compressor
rather than in the
target model's command of the language. The sole exception, Nova~2~Lite
(2/9 significant), is underpowered rather than truly unaffected: it barely
uses the
uncompressed context either (mean full-minus-no-context margin only 16~pp in
English), so its normalized denominators are small and its confidence
intervals correspondingly wide.

\paragraph{Refusals as a compression-induced failure mode.} Claude
frequently declines to answer rather than guessing; we score refusals as
incorrect. With the full passage refusals are negligible ($\le$0.7\% in
every language), but they rise with degradation: up to 11\% of
truncated/randomly-pruned Chinese calls, and 9--31\% of no-context calls
depending on language (27.7\% in English, 31.3\% in Chinese). This has two
consequences. In deployment, aggressive compression of non-Latin text
can convert answers into user-visible refusals, a failure mode invisible
in English-centric evaluation and absent in GPT-5.4-mini, which answers
regardless. For our analysis, refusal deflates Claude's no-context anchor
in all languages, including English, so its net effect on the
cross-language gap estimates has no single direction; we flag it as a
behavioral confound bundled into what ``degraded context'' means for this
model family, and note that the GPT results, which are refusal-free,
reproduce every headline pattern.

\section{Long-Context Stress Test}
\label{sec:longctx}

\begin{figure}[t]
  \centering
  \includegraphics[width=\textwidth]{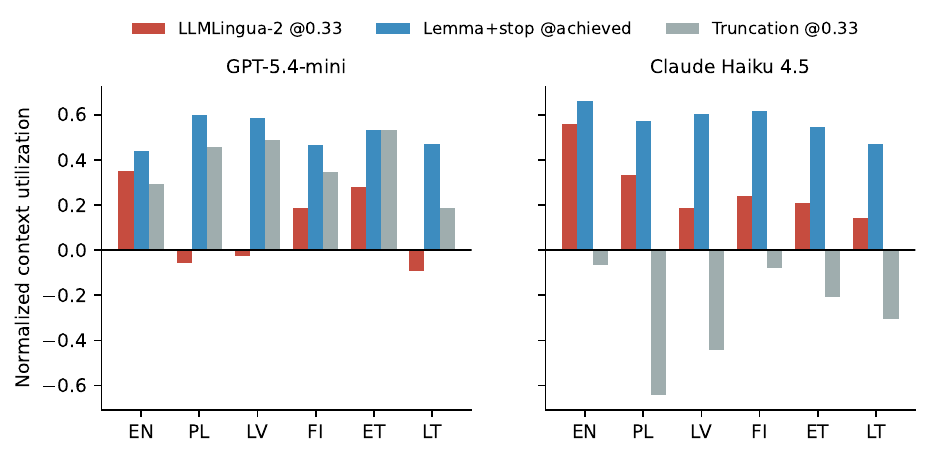}
  \caption{Long-context arm (150 items; target passage among 7 distractors,
  question-agnostic compression). LLMLingua-2 at $\keeprate=0.33$ keeps
  English context useful but drives non-English contexts to (or below)
  no-context utility for GPT; deterministic lemma+stopword compression at its
  natural budget preserves 44--66\% everywhere.}
  \label{fig:longctx}
\end{figure}

When the relevant passage must survive compression among distractors
(Figure~\ref{fig:longctx}), the pattern becomes stronger. With GPT, compressed
Lithuanian, Latvian, and Polish contexts perform at or below the no-context
anchor ($U \le 0$); the compressed context gives no benefit over sending no
context, while the user still pays for its tokens.
Claude replicates the ordering (EN 0.56 vs.\ 0.14--0.33 elsewhere).

\paragraph{Task-dependence.} On MultiEURLEX topic classification the result
is the opposite: compression is nearly free in all languages down to
20$\times$, and a title-only heuristic (5\% budget) matches or beats the full
document (micro-F1 0.47--0.51 vs.\ 0.45--0.48; full table in
Appendix~\ref{app:eurlex}). The cross-lingual penalty is thus a property of
tasks whose answers depend on distributed content rather than of
compression itself; for surface-signal tasks, positional heuristics beat
learned compression at a fraction of the cost.

\section{Translation Arbitrage}
\label{sec:arbitrage}

Translating the context to English (NLLB-200-distilled-600M
\citep{nllb2022}; CC~BY-NC, like XProvence) before compression removes the
token premium at the source:
translation alone saves 35--44\% of tokens at a 5--8\,pp quality cost, and
translate-then-compress delivers, at $\sim$0.18$\times$ the native token
cost, accuracy that matches or beats native-language compression at
0.33$\times$ cost in three of five languages (LT $+10$\,pp, FI $+6.6$,
ET $+5$; Table~\ref{tab:arbitrage}). With $n=60$ items per language we read
this arm as directional rather than definitive. Compression of the
translated
text costs only as much as it does for native English, consistent with the
gap following the surface language rather than the semantics. Concurrent work
\citep{colak2026arbitrage} operationalizes the same economics as an
edge-side translate-and-rewrite middleware for code agents; our controlled
parallel-data results explain when and why such pipelines beat
native-language compression.

\begin{table}[t]
\centering\small
\caption{Translation arbitrage (GPT-5.4-mini, $n=60$ items per language).
Accuracy (\%) by context path; cost = context tokens relative to the
native-language context (o200k). Native LL2@0.33 costs 0.33 by construction.}
\label{tab:arbitrage}
\begin{tabular}{lcccccc}
\toprule
Language & Native full & NLLB full & Native LL2@0.33 & NLLB+LL2 & Cost NLLB & Cost NLLB+LL2 \\
\midrule
PL & 93.3 & 91.7 & 76.7 & 71.7 & 0.60 & 0.18 \\
LV & 91.7 & 83.3 & 71.7 & 70.0 & 0.56 & 0.17 \\
FI & 96.7 & 91.7 & 71.7 & 78.3 & 0.63 & 0.19 \\
ET & 96.7 & 86.7 & 63.3 & 68.3 & 0.65 & 0.20 \\
LT & 98.3 & 91.7 & 68.3 & 78.3 & 0.57 & 0.17 \\
\bottomrule
\end{tabular}
\end{table}

This arm has two caveats. First, our passages descend from FLORES (translated from
English), so round-trip translationese may flatter this pipeline;
NLLB-200 was itself developed around FLORES-200, so its translation quality
here is likely an upper bound. A
native-source replication is future work. Second, the pipeline adds
translation latency ($\sim$1--2\,s per passage for the 600M model on an
M-series laptop GPU, sentence-level greedy decoding; LLMLingua-2 itself adds
$\sim$0.1--0.3\,s per passage). For batch or cache-warm workloads this is
negligible; for interactive first-token latency it may not be.

\section{Discussion and Limitations}
\label{sec:discussion}

\paragraph{Why does the learned compressor fail?} Diagnostics point toward
grammatical cohesion rather than fact deletion. At $\keeprate=0.33$,
number-bearing tokens are retained more often in Lithuanian (0.70),
Latvian (0.78), and Polish (0.70) than in English (0.61), yet quality falls
far more in those languages; conversely, capitalized-word retention (a
named-entity proxy, defined for bicameral scripts only) is highest in
English (0.86 vs.\ 0.72--0.79). The
compressor does not simply delete facts from non-English text; it appears
to disrupt the surface carriers of grammatical structure, which differ by
language. The supervision language predicts which compressors show the gap.
The typology of the target language shapes the form the damage takes.
In the Baltic, Slavic, and Finnic
languages, thematic roles live in case endings and agreement morphology
rather than word order, so dropping ``low-information'' function morphology
is not free; a case-marking minimal-pair probe to test this mechanism
directly is under native-speaker validation (Appendix~\ref{app:probe}).
Chinese (isolating, case-less, and the worst-hit language) requires a
different account: most Chinese words are multi-character, so
character-level deletion yields different words or non-words, and
grammatical particles (the object marker \emph{ba}, passive marker
\emph{bei}, aspect marker \emph{le}, subordinator \emph{de})
are exactly the high-frequency ``function tokens'' an English-trained
classifier learns to discard; the segmentation and budget-control failures
documented above compound this. The probe tests the inflectional mechanism
only; a Chinese diagnostic is future work.

\paragraph{Practical recommendations.} The audit supports a small set of
deployment rules. (1)~\emph{Gate on task type}: for surface-signal tasks
(topic routing, classification) compress aggressively in any language;
positional heuristics at a 5\% budget match the full document
(Section~\ref{sec:longctx}). For answer-bearing tasks, the rules below
apply. (2)~\emph{If the query is known at compression time}, a
multilingually trained query-aware pruner is the only method in our set
that is safe in all ten languages ($U = 0.68$--$0.94$ at $\approx$2$\times$
compression), subject to its non-commercial license and to a
per-language calibration check: the v2 release returns empty
contexts for 92\% of Chinese inputs at its aggressive threshold, with no
error or warning.
(3)~\emph{For
query-agnostic learned compression outside English, stay at mild budgets}:
at a $U \ge 0.8$ fidelity bar, no English-supervised learned compressor
qualifies below $\keeprate = 0.75$ in any non-English language on
either primary model; for Chinese, no tested rate qualifies at all.
(4)~\emph{At deep budgets outside English, prefer deterministic methods}:
lemmatization+stopword removal retains $U = 0.45$--$0.83$ at its natural
$\approx$0.72 budget where morphological resources exist, and TF-IDF
degrades smoothly without a cross-lingual cliff. (5)~\emph{Never apply
whitespace-based or English-supervised compressors to Chinese} (no-ops
without any warning, or $U \le 0$). (6)~\emph{Monitor achieved rates per
language in
production}: the failure modes we document (no-ops,
over-compression, empty outputs, budget drift) produce no error or warning,
but all of them are visible as a
divergence between
requested and achieved rate, at negligible logging cost. Two points about
economics also apply: compressing a shared, prefix-cached context saves less
than naive token arithmetic suggests, and query-aware pruning produces a
unique context per query, forfeiting prefix-cache reuse entirely;
task-agnostic compressions, by contrast, can be cached per document.

\paragraph{Limitations.} Ten languages and five scripts is still a small
slice of the world's languages (six of ten are Indo-European; no non-Uralic
agglutinative language such as Turkish), and the auxiliary arms
(long-context, MultiEURLEX, arbitrage) cover only the six core European
languages. FLORES-derived passages carry contamination and
translationese risks (mitigated by no-context anchoring and normalized
scoring, not eliminated); note the direction of the translationese bias:
our non-English passages are translations from English and thus
syntactically closer to English than natively authored text, so the
measured transfer gap is plausibly a lower bound for native text.
XProvence's query-aware setting is easier than
task-agnostic compression, so its null gap is a supervision-language
contrast rather than a like-for-like quality comparison; refusal behavior differs
across target models (Section~\ref{sec:results}). Statistically, items
cluster within 166 passages while our primary bootstrap resamples items; a
passage-level cluster bootstrap leaves every headline significance call
unchanged (0 of 177 comparisons flip on GPT-5.4-mini, median CI width
ratio 1.02; 3 borderline flips on Claude, all toward significance
and none involving LLMLingua-2). We apply no multiplicity correction
across the several hundred reported
intervals, which favors detecting effects; this is another reason we rest
no conclusion on any single interval and emphasize replication across
models, methods, and rates. Contexts here are short ($\le$3k tokens);
production RAG contexts are often 10--100$\times$ longer, and instructions
were held constant in English rather than localized.

\section{Conclusion}
\label{sec:conclusion}
English-centric evaluation of prompt compressors clearly overstates their
multilingual usefulness. Safe compression budgets outside English are roughly
half those in English for every English-supervised learned compressor we
audited (across three backbones and a production system), while
trivial deterministic baselines transfer with little or no gap and match or
beat the learned methods at deep budgets outside English. This failure can
be fixed: a pruner trained on natively multilingual data (XProvence~v1)
shows no gap at all, which places the problem in the compression
supervision data rather than in the architecture. Its translation-trained
successor (v2), however, shows that the fix is not stable; it over-prunes
Chinese to empty contexts at aggressive settings. Practitioners compressing non-English contexts
today should prefer
budget-matched deterministic methods, multilingually supervised compressors
where available, or translate-then-compress pipelines\footnote{Note that
the two off-the-shelf remedies we test are non-commercial checkpoints:
XProvence is CC~BY-NC-ND~4.0 and NLLB-200 is CC~BY-NC~4.0. Commercial
deployments need a licensed multilingual pruner or MT system, or must fall
back on the deterministic methods.};
researchers should report cross-lingual rate--utility curves instead of
single-rate English scores.

\paragraph{Reproducibility.} All code, cached compressions
(25{,}000+ compressed contexts with achieved budgets), and raw model outputs
(178{,}000+ evaluation records in the main arms alone) are released at
\url{https://github.com/MantasLukauskas/lost-in-compression}.
Pipeline details in Appendix~\ref{app:repro}.

\bibliographystyle{plainnat}
\FloatBarrier
\bibliography{references}

\appendix
\section{Token Premium Measurements}
\label{app:premium}
\begin{figure}[h]
  \centering
  \includegraphics[width=0.8\textwidth]{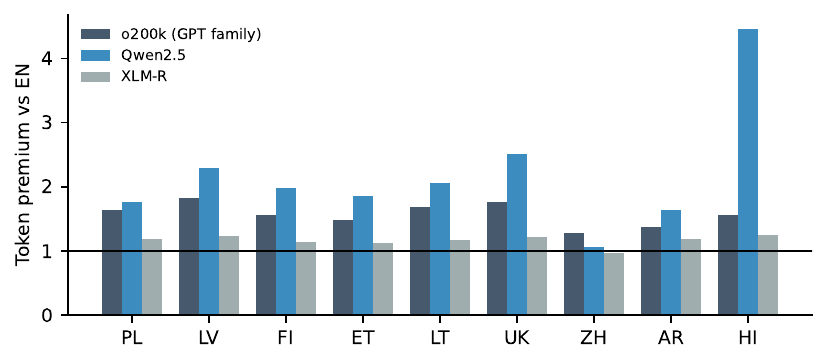}
  \caption{Corpus-level token premium vs.\ English on parallel Belebele
  passages, by tokenizer. XLM-R (the compressor's internal tokenizer) prices
  the ten languages nearly equally (0.97--1.25): the compressor receives
  balanced input lengths, so input length does not explain the transfer
  failure. Hindi's premium
  under Qwen2.5 reaches 4.45$\times$.}
  \label{fig:premium}
\end{figure}

Figure~\ref{fig:premium} reports the corpus-level token premium relative to
English, measured on the 300 parallel Belebele passages under each of the
four tokenizers considered in the study. The compressor-internal XLM-R
tokenizer prices all ten languages nearly equally, which rules out input
length as the driver of the transfer gap.

\FloatBarrier
\section{Full Results Tables}
\label{app:full}
Tables~\ref{tab:acc-gpt} and~\ref{tab:acc-claude} give raw accuracies for
every condition, requested rate, and language on the two primary target
models; Table~\ref{tab:gapsfull} gives the LLMLingua-2 transfer gaps with
bootstrap confidence intervals, and Table~\ref{tab:allmodels} the
cross-model summary behind Figure~\ref{fig:models}.

\begin{table}[h]
\centering\scriptsize
\caption{Accuracy (\%) by condition, requested rate, and language;
GPT-5.4-mini, $n=300$ paired items, temperature 0, zero parse failures.
Truncation and random pruning are matched per item to LLMLingua-2's achieved
o200k budget at the corresponding rate. Lemma+stopword is undefined for
ZH/AR/HI (no simplemma support).}
\label{tab:acc-gpt}
\resizebox{\textwidth}{!}{\begin{tabular}{llcccccccccc}
\toprule
Condition & Rate & EN & PL & LV & FI & ET & LT & UK & ZH & AR & HI \\
\midrule
Full context & -- & 97.0 & 92.3 & 94.3 & 91.0 & 93.0 & 94.0 & 95.0 & 95.0 & 95.0 & 86.7 \\
No context & -- & 70.3 & 66.3 & 63.0 & 64.7 & 63.0 & 66.7 & 69.3 & 63.3 & 69.3 & 58.7 \\
Lemma+stopword & $\sim$0.72 & 82.3 & 83.3 & 86.0 & 78.7 & 84.0 & 82.7 & 90.7 & -- & -- & -- \\
\midrule
LLMLingua-2 (XLM-R) & 0.75 & 93.7 & 90.0 & 88.0 & 88.0 & 87.0 & 89.0 & 89.7 & 83.0 & 90.0 & 80.3 \\
 & 0.50 & 90.0 & 81.3 & 82.3 & 77.0 & 78.7 & 77.3 & 83.0 & 69.7 & 82.7 & 70.3 \\
 & 0.33 & 85.7 & 77.0 & 73.7 & 72.0 & 69.0 & 69.3 & 72.0 & 62.3 & 73.3 & 68.0 \\
\midrule
LLMLingua-2 (mBERT) & 0.75 & 94.0 & 91.0 & 88.0 & 85.7 & 87.7 & 88.0 & 90.3 & 88.0 & 88.3 & 80.7 \\
 & 0.50 & 90.3 & 83.0 & 79.3 & 76.7 & 75.7 & 77.0 & 79.7 & 76.0 & 75.7 & 70.7 \\
 & 0.33 & 80.7 & 76.3 & 73.3 & 68.0 & 68.7 & 71.0 & 74.0 & 70.0 & 71.7 & 66.0 \\
\midrule
TF-IDF & 0.75 & 85.0 & 84.0 & 81.3 & 82.0 & 77.3 & 83.3 & 83.0 & 80.3 & 80.0 & 73.0 \\
 & 0.50 & 72.7 & 73.0 & 75.3 & 74.0 & 72.7 & 75.0 & 77.7 & 71.3 & 71.3 & 66.0 \\
 & 0.33 & 71.3 & 70.3 & 68.7 & 69.7 & 74.3 & 71.0 & 72.0 & 66.7 & 70.3 & 64.0 \\
\midrule
Truncation & 0.75 & 85.0 & 85.3 & 85.0 & 82.3 & 83.0 & 86.0 & 85.3 & 80.3 & 85.0 & 76.0 \\
 & 0.50 & 77.0 & 74.0 & 76.7 & 75.3 & 73.7 & 76.3 & 77.0 & 71.3 & 77.3 & 68.0 \\
 & 0.33 & 73.7 & 70.0 & 71.0 & 69.0 & 65.7 & 69.0 & 67.3 & 66.0 & 68.3 & 62.3 \\
\midrule
Random pruning & 0.75 & 88.3 & 84.7 & 83.7 & 81.0 & 83.3 & 83.3 & 87.3 & 79.3 & 87.3 & 77.0 \\
 & 0.50 & 74.3 & 75.7 & 73.0 & 70.0 & 69.0 & 73.3 & 78.0 & 67.0 & 73.3 & 65.0 \\
 & 0.33 & 72.3 & 71.0 & 66.7 & 66.0 & 65.3 & 67.7 & 70.7 & 66.7 & 68.3 & 62.0 \\
\midrule
Kompress-v2 & 0.50 & 88.0 & 75.3 & 76.7 & 75.3 & 73.0 & 76.7 & 80.7 & 91.7 & 75.0 & 71.0 \\
 & 0.33 & 80.3 & 72.0 & 69.7 & 67.0 & 67.3 & 69.3 & 72.3 & 88.3 & 68.7 & 64.7 \\
\midrule
XProvence v1 (thr.) & 0.10 & 96.7 & 92.0 & 94.0 & 92.7 & 91.7 & 94.0 & 94.0 & 95.0 & 94.3 & 87.0 \\
 & 0.50 & 91.0 & 89.0 & 87.0 & 86.0 & 86.3 & 88.7 & 89.7 & 85.0 & 90.3 & 85.0 \\
\midrule
XProvence v2 (thr.) & 0.10 & 93.7 & 92.7 & 91.7 & 92.0 & 91.3 & 92.0 & 93.3 & 94.7 & 93.3 & 85.7 \\
 & 0.50 & 85.0 & 80.7 & 78.0 & 76.3 & 77.7 & 79.3 & 82.0 & 64.3 & 76.3 & 66.3 \\
\bottomrule
\end{tabular}}
\end{table}

\begin{table}[h]
\centering\scriptsize
\caption{Accuracy (\%) by condition, requested rate, and language;
Claude Haiku 4.5, $n=300$ paired items. Refusals ($\sim$5\% of calls,
concentrated in degraded conditions and non-Latin scripts) scored as
incorrect.}
\label{tab:acc-claude}
\resizebox{\textwidth}{!}{\begin{tabular}{llcccccccccc}
\toprule
Condition & Rate & EN & PL & LV & FI & ET & LT & UK & ZH & AR & HI \\
\midrule
Full context & -- & 95.0 & 93.7 & 92.7 & 95.3 & 91.7 & 92.7 & 94.3 & 95.0 & 93.0 & 83.7 \\
No context & -- & 53.0 & 64.0 & 57.3 & 49.7 & 55.0 & 57.0 & 61.7 & 50.7 & 58.0 & 43.3 \\
Lemma+stopword & $\sim$0.72 & 81.3 & 80.7 & 83.0 & 74.7 & 78.3 & 80.3 & 86.7 & -- & -- & -- \\
\midrule
LLMLingua-2 (XLM-R) & 0.75 & 93.0 & 89.0 & 88.3 & 87.7 & 86.0 & 86.3 & 91.3 & 83.0 & 88.7 & 80.3 \\
 & 0.50 & 87.7 & 80.0 & 76.7 & 76.3 & 74.7 & 75.3 & 79.0 & 71.0 & 82.3 & 71.7 \\
 & 0.33 & 79.0 & 69.3 & 70.7 & 66.0 & 64.3 & 65.7 & 70.0 & 62.0 & 68.3 & 64.7 \\
\midrule
LLMLingua-2 (mBERT) & 0.75 & 94.0 & 90.0 & 85.7 & 88.3 & 86.0 & 86.0 & 87.7 & 88.3 & 90.0 & 79.3 \\
 & 0.50 & 84.0 & 80.7 & 74.7 & 75.3 & 71.7 & 74.0 & 77.3 & 74.7 & 74.3 & 70.3 \\
 & 0.33 & 76.7 & 71.0 & 66.0 & 67.0 & 67.3 & 64.3 & 70.0 & 66.0 & 64.3 & 63.7 \\
\midrule
TF-IDF & 0.75 & 83.3 & 81.0 & 78.0 & 82.0 & 76.3 & 80.0 & 81.7 & 79.3 & 81.3 & 73.7 \\
 & 0.50 & 67.3 & 69.3 & 70.3 & 73.0 & 66.3 & 71.7 & 72.7 & 68.0 & 69.7 & 66.7 \\
 & 0.33 & 66.7 & 66.3 & 65.7 & 67.7 & 66.0 & 68.7 & 67.0 & 65.7 & 66.0 & 62.0 \\
\midrule
Truncation & 0.75 & 83.0 & 83.3 & 82.0 & 83.0 & 82.3 & 81.0 & 80.3 & 80.7 & 83.7 & 75.0 \\
 & 0.50 & 72.0 & 72.3 & 70.0 & 71.0 & 70.0 & 72.0 & 71.0 & 68.0 & 72.7 & 64.3 \\
 & 0.33 & 63.3 & 62.0 & 59.3 & 58.3 & 58.0 & 58.3 & 58.7 & 59.7 & 62.3 & 54.3 \\
\midrule
Random pruning & 0.75 & 86.0 & 85.0 & 82.0 & 82.7 & 76.3 & 81.0 & 82.0 & 77.0 & 86.3 & 74.0 \\
 & 0.50 & 73.0 & 69.0 & 68.0 & 69.0 & 66.0 & 72.0 & 71.7 & 66.3 & 69.7 & 65.3 \\
 & 0.33 & 64.0 & 60.3 & 58.3 & 60.0 & 57.7 & 59.3 & 62.3 & 59.7 & 57.0 & 52.0 \\
\midrule
Kompress-v2 & 0.50 & 87.0 & 74.3 & 70.3 & 73.0 & 69.7 & 71.7 & 74.0 & 92.0 & 70.3 & 66.0 \\
 & 0.33 & 72.3 & 61.7 & 62.3 & 63.3 & 60.3 & 62.7 & 63.3 & 88.0 & 63.3 & 61.7 \\
\midrule
XProvence v1 (thr.) & 0.10 & 95.0 & 93.7 & 92.7 & 95.3 & 91.7 & 92.3 & 93.7 & 94.0 & 93.0 & 84.0 \\
 & 0.50 & 87.0 & 84.7 & 84.3 & 84.7 & 79.7 & 84.3 & 87.0 & 69.7 & 86.7 & 82.0 \\
\midrule
XProvence v2 (thr.) & 0.10 & 93.3 & 92.3 & 90.3 & 91.7 & 89.3 & 90.0 & 92.0 & 94.0 & 92.0 & 83.3 \\
 & 0.50 & 68.0 & 68.0 & 64.3 & 63.7 & 59.7 & 64.3 & 66.0 & 23.7 & 55.7 & 49.0 \\
\bottomrule
\end{tabular}}
\end{table}

\begin{table}[h]
\centering\scriptsize
\caption{LLMLingua-2 transfer gap $\tau$ vs.\ EN with 95\% item-level paired
bootstrap CIs (2{,}000 resamples). Bold: CI excludes zero.}
\label{tab:gapsfull}
\begin{tabular}{llccc}
\toprule
Model & Language & $\keeprate=0.75$ & $\keeprate=0.5$ & $\keeprate=0.33$ \\
\midrule
GPT-5.4-mini & PL & -0.04 [-0.17, +0.08] & +0.16 [-0.02, +0.35] & +0.17 [-0.04, +0.37] \\
 & LV & +0.08 [-0.04, +0.19] & +0.12 [-0.03, +0.27] & \textbf{+0.23 [+0.05, +0.44]} \\
 & FI & -0.01 [-0.16, +0.13] & \textbf{+0.27 [+0.08, +0.46]} & \textbf{+0.30 [+0.09, +0.51]} \\
 & ET & +0.07 [-0.05, +0.20] & \textbf{+0.21 [+0.04, +0.39]} & \textbf{+0.38 [+0.18, +0.60]} \\
 & LT & +0.06 [-0.06, +0.18] & \textbf{+0.35 [+0.17, +0.52]} & \textbf{+0.48 [+0.26, +0.73]} \\
 & UK & +0.08 [-0.04, +0.21] & \textbf{+0.20 [+0.04, +0.38]} & \textbf{+0.47 [+0.26, +0.72]} \\
 & ZH & \textbf{+0.25 [+0.11, +0.40]} & \textbf{+0.54 [+0.34, +0.76]} & \textbf{+0.61 [+0.39, +0.85]} \\
 & AR & +0.07 [-0.07, +0.20] & \textbf{+0.22 [+0.03, +0.41]} & \textbf{+0.42 [+0.22, +0.65]} \\
 & HI & +0.10 [-0.03, +0.23] & \textbf{+0.32 [+0.14, +0.52]} & \textbf{+0.24 [+0.06, +0.44]} \\
\midrule
Claude Haiku 4.5 & PL & \textbf{+0.11 [+0.02, +0.21]} & \textbf{+0.29 [+0.14, +0.43]} & \textbf{+0.44 [+0.27, +0.63]} \\
 & LV & +0.07 [-0.01, +0.16] & \textbf{+0.28 [+0.15, +0.42]} & \textbf{+0.24 [+0.11, +0.38]} \\
 & FI & \textbf{+0.12 [+0.05, +0.20]} & \textbf{+0.24 [+0.14, +0.35]} & \textbf{+0.26 [+0.14, +0.39]} \\
 & ET & \textbf{+0.11 [+0.02, +0.20]} & \textbf{+0.29 [+0.17, +0.42]} & \textbf{+0.36 [+0.22, +0.53]} \\
 & LT & \textbf{+0.13 [+0.04, +0.22]} & \textbf{+0.31 [+0.18, +0.45]} & \textbf{+0.38 [+0.24, +0.54]} \\
 & UK & +0.04 [-0.04, +0.13] & \textbf{+0.29 [+0.18, +0.42]} & \textbf{+0.36 [+0.21, +0.54]} \\
 & ZH & \textbf{+0.22 [+0.12, +0.33]} & \textbf{+0.37 [+0.25, +0.48]} & \textbf{+0.36 [+0.21, +0.52]} \\
 & AR & +0.08 [-0.01, +0.16] & \textbf{+0.13 [+0.01, +0.25]} & \textbf{+0.32 [+0.18, +0.48]} \\
 & HI & +0.04 [-0.04, +0.12] & \textbf{+0.12 [+0.01, +0.24]} & +0.09 [-0.04, +0.23] \\
\bottomrule
\end{tabular}
\end{table}

\begin{table}[h]
\centering\small
\caption{Cross-model summary: LLMLingua-2 mean transfer gap over the nine
non-English languages (Chinese gap in parentheses) and number of languages
with a significant gap (95\% bootstrap CI excludes zero), per target model.
Primary models use $n=300$; replication models use $n=150$ on a reduced grid,
so deep-budget ($\keeprate=0.33$) significance counts are noisier as
accuracies compress toward the no-context floor. Chinese is the
worst-affected language for every model.}
\label{tab:allmodels}
\begin{tabular}{lcccc}
\toprule
\multirow{2}{*}{Model} & \multicolumn{2}{c}{$\keeprate=0.5$} & \multicolumn{2}{c}{$\keeprate=0.33$} \\
\cmidrule(lr){2-3}\cmidrule(lr){4-5}
 & mean gap (ZH) & \#sig & mean gap (ZH) & \#sig \\
\midrule
GPT-5.4-mini & $+0.27$ ($+0.54$) & 7/9 & $+0.37$ ($+0.61$) & 8/9 \\
Claude Haiku 4.5 & $+0.26$ ($+0.37$) & 9/9 & $+0.31$ ($+0.36$) & 8/9 \\
GPT-5.6-luna & $+0.35$ ($+0.65$) & 9/9 & $+0.30$ ($+0.56$) & 4/9 \\
Gemini 3.5 Flash & $+0.33$ ($+0.64$) & 6/9 & $+0.43$ ($+0.61$) & 9/9 \\
Llama 4 Maverick & $+0.31$ ($+0.50$) & 7/9 & $+0.21$ ($+0.20$) & 2/9 \\
Mistral Medium 3.5 & $+0.40$ ($+0.61$) & 9/9 & $+0.27$ ($+0.55$) & 2/9 \\
DeepSeek V4 Flash & $+0.35$ ($+0.82$) & 6/9 & $+0.11$ ($+0.32$) & 0/9 \\
Kimi K2.6 & $+0.34$ ($+0.72$) & 7/9 & $+0.35$ ($+0.53$) & 5/9 \\
MiniMax M3 & $+0.37$ ($+0.70$) & 8/9 & $+0.39$ ($+0.85$) & 4/9 \\
Nova 2 Lite & $+0.28$ ($+0.60$) & 2/9 & $-0.13$ ($+0.23$) & 0/9 \\
Qwen 3.7 Plus & $+0.40$ ($+0.72$) & 9/9 & $+0.34$ ($+0.77$) & 2/9 \\
\bottomrule
\end{tabular}
\end{table}

\FloatBarrier
\section{MultiEURLEX Task-Dependence Results}
\label{app:eurlex}
Table~\ref{tab:eurlex} reports the topic-classification arm on the six core
languages, where compression is nearly free at every rate; this contrast
establishes the task dependence of the transfer gap
(Section~\ref{sec:longctx}).

\begin{table}[h]
\centering\small
\caption{Micro-F1 on level-1 EUROVOC classification (21 classes;
24 parallel documents per language; GPT-5.4-mini). Compression is nearly
free at all rates in all languages, and the title-only heuristic at a 5\%
budget matches or beats the full document; topic classification does not
probe the cross-lingual gap.}
\label{tab:eurlex}
\begin{tabular}{lcccccc}
\toprule
Condition & EN & PL & LV & FI & ET & LT \\
\midrule
Full document & 0.46 & 0.45 & 0.46 & 0.45 & 0.48 & 0.46 \\
LLMLingua-2 @0.33 & 0.43 & 0.46 & 0.44 & 0.45 & 0.44 & 0.43 \\
LLMLingua-2 @0.15 & 0.44 & 0.45 & 0.42 & 0.47 & 0.46 & 0.46 \\
LLMLingua-2 @0.05 & 0.46 & 0.44 & 0.44 & 0.43 & 0.35 & 0.39 \\
Truncation @0.05 & 0.43 & 0.43 & 0.45 & 0.47 & 0.45 & 0.44 \\
Title only ($\sim$0.05) & 0.47 & 0.51 & 0.48 & 0.50 & 0.50 & 0.47 \\
No document & 0.09 & 0.09 & 0.09 & 0.09 & 0.09 & 0.09 \\
\bottomrule
\end{tabular}
\end{table}

\FloatBarrier
\section{Case-Marking Probe}
\label{app:probe}
Design: minimal pairs in which thematic roles are carried by case endings
under OVS order in Lithuanian (natural word order variation) but by SVO
word order in English; the critical sentence is embedded among filler
sentences and the MCQ asks who performed the action (options: actor,
patient, both, not stated). We construct 75 items from a closed template
inventory (six nouns $\times$ five transitive verbs, gender-agreeing
options).

\emph{Preliminary} results on these template-constructed,
native-speaker-reviewed items (GPT-5.4-mini,
450 calls): with the full passage, both languages are at ceiling (100\%).
Under LLMLingua-2 at $\keeprate=0.33$, English stays at 98.7\% while
Lithuanian falls to 82.7\%. No model response ever selected the patient,
though we note the question format itself disfavors that response (the
patient appears in the question, in the accusative), so the absence of
role reversals should not be over-read; errors are retreats to ``not
stated'', consistent with compression destroying the cues that license the
role assignment rather than deleting the fact itself. The word-overlap
survival metric (89\% for Lithuanian vs.\ 75\% for English) is coarse
(the English critical sentence has only four word types, so the difference
largely reflects function-word deletion), and a morpheme-level metric
(does the case suffix itself survive?) with per-item conditional analysis
is required before this probe can carry causal weight. An extended version
of the probe (role-counterbalanced items,
shuffled full-proposition options, a role-neutral question, LT~SVO~/
EN-passive controls, and a per-item suffix-survival analysis) is left
for future work.

\section{Reproducibility Details}
\label{app:repro}
Compressors: \texttt{microsoft/\allowbreak llmlingua-2-\allowbreak xlm-roberta-\allowbreak large-\allowbreak meetingbank}
and \texttt{microsoft/\allowbreak llmlingua-2-\allowbreak bert-base-\allowbreak multilingual-\allowbreak cased-\allowbreak meetingbank}
(public checkpoints), MPS backend, \texttt{force\_tokens=[``\textbackslash
n'', ``?'']}. Token accounting: \texttt{o200k\_base} via tiktoken for
budgets; achieved rates recorded per item. TF-IDF baseline: sentence-level
extraction (sentence terminators extended for Han, Arabic, and Devanagari
punctuation; character-level term units for ZH), order-preserving, IDF
estimated per language on the full aligned passage pool. Lemma+stopword
baseline: \texttt{simplemma} + \texttt{stopwordsiso}, punctuation dropped;
defined for the seven languages simplemma supports (not ZH/AR/HI). Controls
seeded with a fixed global seed (20260718). The passage-cluster bootstrap
sensitivity analysis is implemented in
\texttt{analyze\_cluster\_sensitivity.py} (multinomial passage weights,
same seed and resample count). Random deletion operates on
whitespace-delimited units; for Chinese, which lacks whitespace, the
deleted units are therefore multi-word chunks rather than words; the
control remains budget-matched, but its granularity is script-dependent. Kompress-v2:
\texttt{chopratejas/\allowbreak kompress-v2-base} via the \texttt{headroom-ai[ml]}
package's \texttt{KompressCompressor} API with explicit
\texttt{target\_ratio} (its word-based ratio accounting explains the
weaker o200k rate adherence); note the higher-level
\texttt{UniversalCompressor} entry point routes plain text to a no-op
handler and must not be used for auditing. XProvence:
\texttt{naver/\allowbreak xprovence-reranker-bgem3-v1} and
\texttt{...-v2}, thresholds 0.1 and 0.5, the
item question as query, sentence segmentation via spaCy
\texttt{xx\_sent\_ud\_sm}; per-item (query-dependent) compression, so its
achieved rate is measured, not controlled; when the pruner returns an
empty string the prompt carries the literal marker \texttt{(empty)}.
Target models:
\texttt{gpt-5.4-mini}, \texttt{claude-haiku-4-5} (full grid);
\texttt{gpt-5.6-luna}, \texttt{gemini-3.5-flash}, Llama~4~Maverick,
Mistral~Medium~3.5, DeepSeek~V4~Flash, Kimi~K2.6, MiniMax~M3,
Nova~2~Lite, and Qwen~3.7~Plus (reduced grid; gateway-served via an
OpenAI-compatible API). Evaluation: single-letter MCQ prompts, English
instructions, temperature 0; answers parsed with a script-agnostic
letter-extraction rule (last standalone A--D, with an explicit
``answer\,X'' override) because word-boundary regexes fail adjacent to
CJK/Arabic/Devanagari characters and verbose open-weights models emit
chain-of-thought before the letter; paired items identical across
languages; bootstrap resamples items, not calls. Total evaluation calls:
90{,}000 + 10{,}800 (main + mBERT arms) + 24{,}000 (Kompress + XProvence~v1
arm, both primary models) + 12{,}000 (XProvence~v2 arm, both primary
models) + 10{,}800 (long-context) + 2{,}304
(MultiEURLEX) + 600 (arbitrage) + 108{,}000 (nine replication models)
$\approx$ 258{,}000.

\end{document}